\documentclass[superscriptaddress,twocolumn]{revtex4}

\usepackage{microtype}
\usepackage{graphicx}
\usepackage{subfigure}
\usepackage{amsmath}
\usepackage{amssymb}

\usepackage{booktabs} 

\usepackage{hyperref}

\begin{document}

\author{Yongjin Park}
\affiliation{Computer Science and Artificial Intelligence Laboratory, Massachusetts Institute of Technology, Cambridge, MA}
\affiliation{Broad Institute of MIT and Harvard, Cambridge, MA}

\author{Abhishek Sarkar}
\affiliation{Department of Human Genetics, University of Chicago, Chicago, IL}

\author{Khoi Nguyen}
\author{Manolis Kellis}
\affiliation{Computer Science and Artificial Intelligence Laboratory, Massachusetts Institute of Technology, Cambridge, MA}
\affiliation{Broad Institute of MIT and Harvard, Cambridge, MA}

\title{Causal Mediation Analysis Leveraging Multiple Types of Summary Statistics Data}

\begin{abstract}
  Summary statistics of genome-wide association studies (GWAS) teach
  causal relationship between millions of genetic markers and tens and
  thousands of phenotypes. However, underlying biological mechanisms
  are yet to be elucidated.  We can achieve necessary interpretation
  of GWAS in a causal mediation framework, looking to establish a
  sparse set of mediators between genetic and downstream variables,
  but there are several challenges. Unlike existing methods rely on
  strong and unrealistic assumptions, we tackle practical challenges
  within a principled summary-based causal inference framework.  We
  analyzed the proposed methods in extensive simulations generated
  from real-world genetic data. We demonstrated only our approach can
  accurately redeem causal genes, even without knowing actual
  individual-level data, despite the presence of competing non-causal
  trails.
\end{abstract}

\maketitle

\section{Introduction}

Genome-wide association studies (GWAS) identify statistically
significant correlations between genetic and phenotypic variables.  In
the era of Biobank GWAS, phenotypes can be virtually any variables
measurable across millions of individuals in the database, of which
examples include diagnosis codes, routine laboratory test results,
family history of complex disorders, and even socio-economical status.

Significant signals of well-executed GWAS implicate unidirectional
causal relationship from the tagged genomic variants to phenotypes,
not the other way.  In biological information cascade, using GWAS, we
can establish links between the very first (genetics) and the last
(phenotypes) layers, and we normally expect the effect sizes are
typically minuscule; and necessary statistical significance can be
achieved in studies involving at least hundreds of thousands of
individuals.  Nonetheless, a large number of GWAS summary statistics
data are already made publicly available.  Geneticists have already
uncovered more than 24k unique associations between single nucleotide
polymorphism (SNP) markers and complex phenotypes
\cite{MacArthur2017-xg}.

However, a fundamental limitation of GWAS remains in its lack of
interpretability. As it can only suggest positions (SNPs) in the human
genome without providing any mechanistic insights into how these loci
exert their action.  Unlike conventional differential gene expression
analysis, nearly 90\% of significant genetic loci fall non-coding
regions \cite{Edwards2013-ro}; therefore, even knowing a target gene
and relevant regulatory context is already a big challenge in most
post-GWAS analysis.  Obviously, by characterization of gene names and
related pathways beyond a set of genomic locations, we can begin to
understand biological mechanisms to find a suitable entry point of
therapeutics.

We recognize interpretation of GWAS can be improved by solving a
series of causal mediation problems.  The basic idea is to jointly
analyze GWAS data with other types of genetic association statistics
that connect genetic variants (SNPs) to \textit{endo}-phenotypes
located in the middle between genetic and phenotypic layers.  We
transfer knowledge of intermediate genetic regulatory mechanisms to
marginalized GWAS summary data \cite{Claussnitzer2015-uj}.

\begin{eqnarray*}
  \textrm{GWAS:}\quad \textsf{SNP} \to & \overbrace{\cdots}^{\textrm{unknown}} & \to \textsf{disease}\\
  \textrm{eQTL:}\quad \textsf{SNP} \to & \textsf{gene} & \\
  \textrm{mediation:}\quad \textsf{SNP} \to & \textsf{gene} & \to \textsf{disease}
\end{eqnarray*}

Of many possible types of endo-phenotypes, we focus on finding a set
of causal genes that mediate between initiating SNPs and target
phenotypes, such as complex diseases.  We leverage the knowledge of
existing eQTL (expression qualitative trait locus) summary statistics.
In eQTL summary statistics data, we compile effect sizes of genetic
associations of nearly 20k genes with common genetic variants
(SNPs). On each gene, approximately 1k-10k neighboring SNPs are
typically tested within a $\pm$ 1 megabase window
(\textit{cis}-eQTLs).  Since we only investigate mediation of
\textit{cis}-regulatory mechanisms, mediation analysis can be
conducted within a segment of genome.  We break down the whole genome
into 1,703 independent blocks \cite{Berisa2016-cm}.

\begin{figure}[h]
  \centering
  \includegraphics[width=\linewidth]{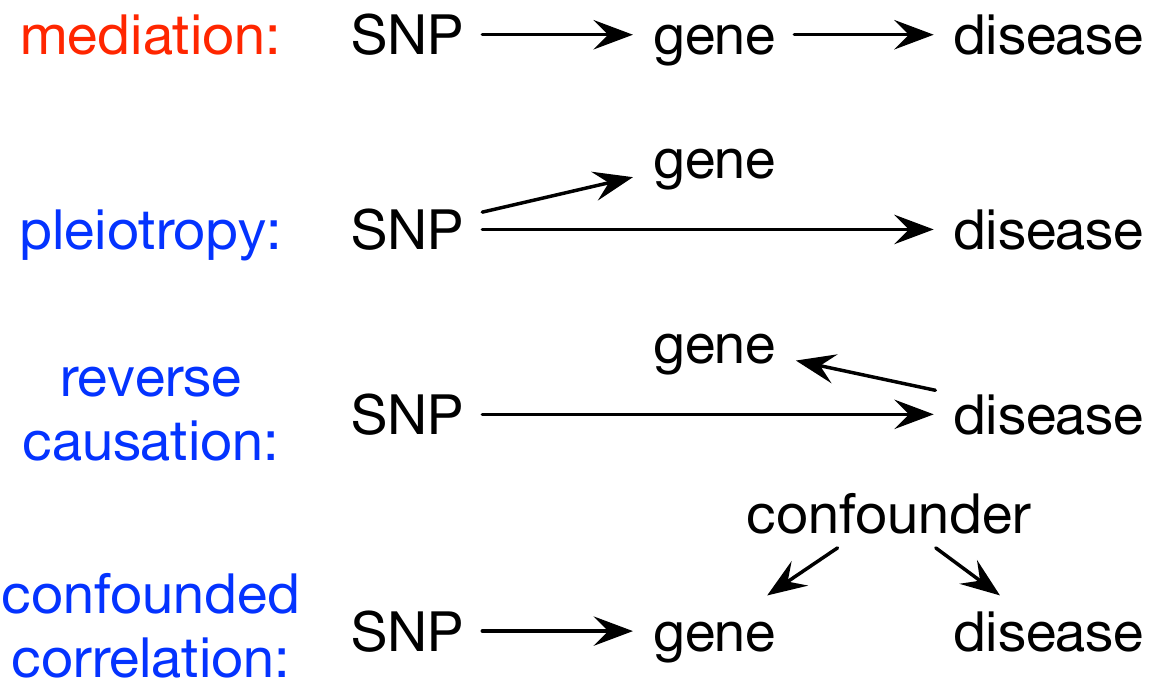}
  \caption{Association between gene and disease can be interpreted in
    multiple ways.}
  \label{fig:causal.trails}
\end{figure}

In ``causal'' mediation analysis, we emphasize that correlation is
never causality because observed gene-disease association /
correlation signals can be interpreted as many different causal
mechanisms (Fig.\ref{fig:causal.trails}).  Of them, we are
particularly interesting in redeeming the mediation effect; only the
mediating gene can causally alter predisposition of the disease.

\paragraph{Our contributions}

In this work, we contribute a general causal inference method for
multivariate mediation analysis, leveraging two types of summary
statistics data--one linking instrumental variables (genetics) to
outcome variables (phenotypes) and the other linking instrumental
variables with mediator variables (endo-phenotypes; genes).

First, we carefully examine the underlying mediation problem in
details, and reveal that a subtle difference can substantially alter
identifiability of the underlying statistical problem.  We claim that
practical issues, such as incomplete knowledge of mediation, polygenic
bias, and uncharacterized confounding effects, should be carefully
controlled; otherwise, association-based methods may pile up wrong
interpretation of GWAS results.

Second, to our knowledge, this work\footnote{This includes our
  previous work that only focus on biological aspects without clear
  exposition of causal inference and machine learning aspects.} is the
first attempt in summary-based mediation analysis to include and test
multiple mediation variables within a single Bayesian framework.  Our
formalism might be summary-based translation of the existing proposal
for fully observed individual-level data \cite{VanderWeele2014-cg},
but we address practical issues hidden underneath the observed summary
statistics.

Third, our approach is built on a principled multivariate model, which
takes into accounts of inherent dependency structure between a large
number of genetic variants.  We resolve the unavoidable issue of
high-dimensional collinearity using sparse Bayesian variable selection
\cite{Mitchell1988-cf}, and demonstrate that our Bayesian approach
yields superior performance in relevant simulations.

Fourth, we propose novel, yet simple, operational steps generally
applicable to summary-based causal inference problems.  We solve a
long-standing problem of confounder correction in genetics data, not
relying on unrealistic simplifications and assumptions.

\paragraph{Related work}

Mendelian Randomization (MR) \cite{Smith2004-mx,Katan2004-yr} resolves
causal directions by using genetic variants as instrumental variables
(IV) in causal inference analysis.  However, MR assumes that entire
proportion of causal effects in the genetic locus on the phenotype are
mediated by the measured intermediate phenotype (e.g. expression of a
given gene in the given cell type)
\cite{Smith2004-mx,Davey_Smith2014-oa}, which asserts there is no
other causal trails exist, and more importantly most MR method only
works on a few IV variants.

Transcriptome-wide association studies (TWAS) aggregates information
of multiple variants to find genes whose regulatory variants have
correlated effect sizes for both gene expression and downstream
phenotypes \cite{Gamazon2015-uw,Gusev2016-oa,Mancuso2017-kp}. However,
TWAS methods are fundamentally limited because they cannot distinguish
between causal mediation, pleiotropy, linkage between causal variants,
and reverse causation, which could lead to inflated false positives.

\section{Causal mediation analysis}

\subsection{A generative model of phenotypic variability mediated by gene expressions}

We model a phenotype vector $\mathbf{y}$ of $n$ individuals as a
function of genotype information measured across $p$ common variants
(SNPs).
\begin{equation}
  \label{eq:qtl.model}
  \mathbf{y} \sim X\boldsymbol{\theta} + \boldsymbol{\epsilon},
  \quad \boldsymbol{\epsilon} \sim \mathcal{N}\!\left(\mathbf{0},\sigma^{2}I\right)
\end{equation}
with the multivariate effect size $\boldsymbol{\theta}$. We assume
irreducible isotropic noise $\epsilon$ fluctuates with some variance
$\sigma^{2}$.  For simplicity, we assume the GWAS trait is
quantitative, $y_{i} \in \mathbb{R}$ and the genotype matrix $X$ is
column-wise standardized with mean zero and unit standard deviation.

Conventional definition of GWAS statistics refers a univariate effect
size (a regression slope of a simple regression (or log-odds ratio) in
case-control studies) measured on each genetic variant.  In summary
data, we have a vector of $p$ summary statistics, effect size
$\hat{\theta}_{j}$ and corresponding variance $\hat{\sigma}^{2}_{j}$
for each SNP $j\in[p]$.
\begin{eqnarray}
  \label{eq:summary.theta}
  \hat{\theta}_{j} = \frac{\mathbf{x}_{j}^{\top}\mathbf{y}}{\mathbf{x}_{j}^{\top}\mathbf{x}_{j}} \quad\textrm{and}\quad
  \hat{\sigma}_{j}^{2} = \frac{(\mathbf{y} -\mathbf{x}_{j}\hat{\theta}_{j})^{\top}(\mathbf{y} - \mathbf{x}_{j}\hat{\theta}_{j})}{n\mathbf{x}_{j}^{\top}\mathbf{x}_{j}}.
\end{eqnarray}
However, due to linkage disequilibrium (LD; correlations between
neighboring SNPs), an effect size measured on each single variant
contains contributions from the neighboring SNPs.

Likewise, expression profiles of $K$ genes are generated by the same
type of models on the shared genotype matrix $X$.  For each gene
$k\in[K]$, we define a generative model of gene expression
$\mathbf{m}_{k}$:
\begin{equation}
  \label{eq:eqtl.model}
  \mathbf{m}_{k} = X\boldsymbol{\alpha}_{k} +
  \boldsymbol{\delta}_{k}, \quad
  \boldsymbol{\delta}_{k} \sim \mathcal{N}\!\left(\mathbf{0}, \tau_{k}^{2}I\right) 
\end{equation}
where multivariate eQTL effect size vector $\boldsymbol{\alpha}_{k}$
exerts an action on each gene $k$, but there is a measurement error
$\delta$ with non-genetic variance $\tau_{k}$.

\begin{figure}[h]
  \centering
  \includegraphics[width=\linewidth]{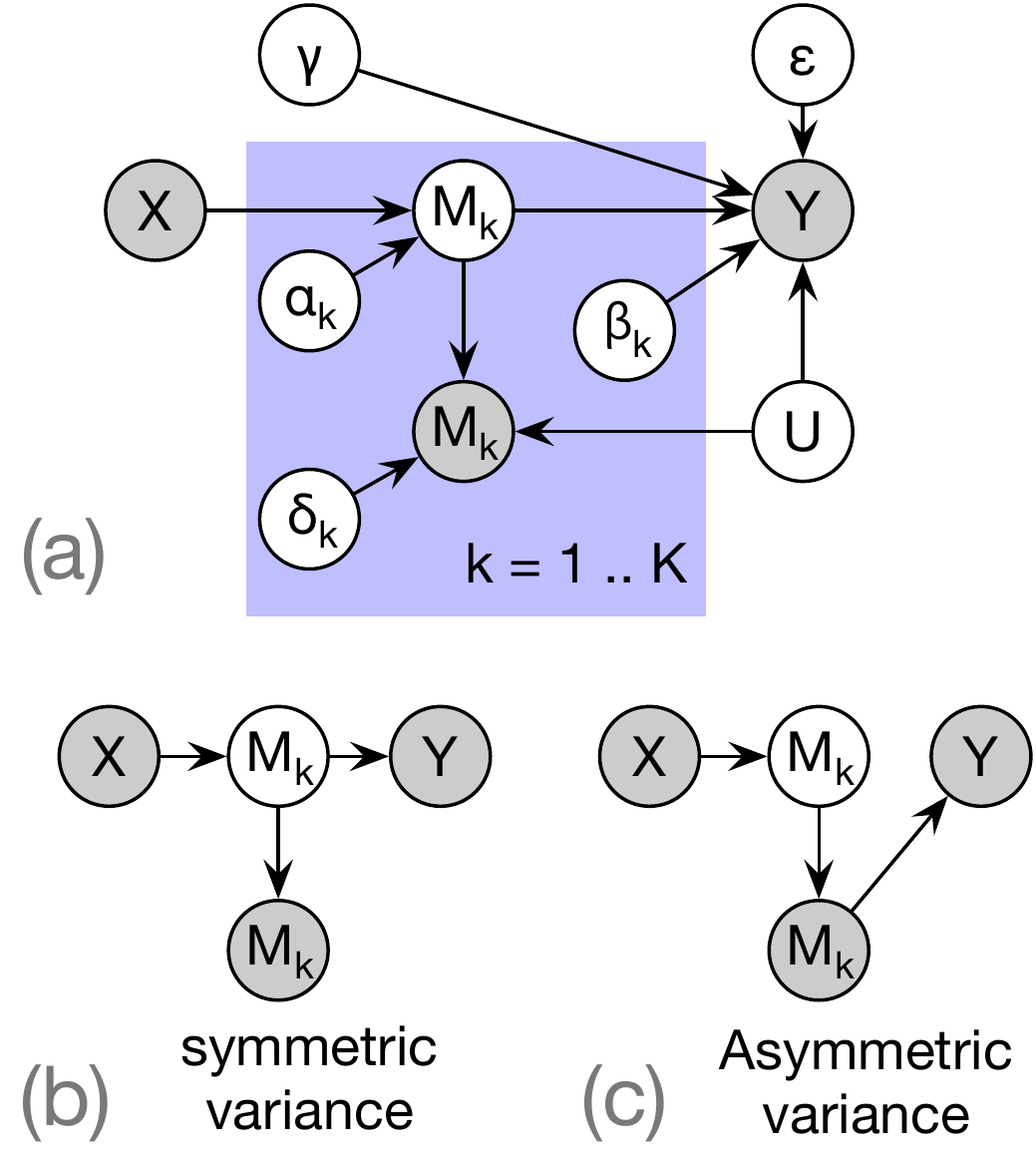}
  \caption{{\bf (a)} A graphical model for the full generative scheme
    with $K$ genes.  $X$: genotype matrix; $M$ (unfilled): genuine
    mediation effect; $M$ (filled): observed mediation variable; $Y$:
    phenotype measurement; $U$: unmeasured confounding variable;
    $\alpha$: eQTL effect size; $\beta$: mediation effect size;
    $\gamma$: unmediated effect size; $\delta$: non-genetic components
    in gene expression variation; $\epsilon$: non-genetic components
    in phenotypic variation.  {\bf (b)} The symmetric variance model
    (see the text). {\bf (c)} The asymmetric variance model (see the
    text).}
  \label{fig:pgm.mediation}
\end{figure}

\subsection{Two types of mediation models}

Before we present methods and algorithms, we digress to dissect
identifiability issues with intuitive examples.

Total genetic effect $X\boldsymbol{\theta}$ on the phenotypic
variation (Eq.\ref{eq:qtl.model}) decomposes into two components,
mediated from the causal genes $X\boldsymbol{\alpha}_{k} \beta_{k}$
with mediation effect size $\beta_{k}$ and unmediated effects
$X\boldsymbol{\gamma}$ with some coefficients $\boldsymbol{\gamma}$.
In other words,
$X\boldsymbol{\theta} = X(\sum_{k}\boldsymbol{\alpha}_{k} \beta_{k} +
\boldsymbol{\gamma})$.

Overall, this generative scheme (Fig.\ref{fig:pgm.mediation}a) is
generally acceptable to most of genetics and epidemiology research
community, but we demonstrate a subtle, yet critical difference about
when we actually measure the mediation profiles with non-genetic
stochasticity $\delta$ (Fig.\ref{fig:pgm.mediation}b versus c) can
make impact on identifiability.  In the former model (b), we assume
that the non-genetic components $\delta$ are mostly attributable to
technical covariates, and estimable through control genes in eQTL
analysis \cite{Gagnon-Bartsch2012-jc,Risso2014-st}, whereas in the
latter case (c), non-genetic signals $\delta$ first biologically
incorporate and become transmitted to downstream phenotypic variation.
We term them \textit{symmetric} and \textit{asymmetric} variance
models, respectively, as they result in different variance structures
in the following model identification steps.

\paragraph{Symmetric variance model}

For intuitive explanation, without loss of generality, suppose we only
have a single gene and a single genetic variant in the model.  If we
assume stochasticity was infused after the mediation, we generate the
phenotype by
\[
  \mathbf{y} = \mathbf{x}\alpha\beta + \mathbf{x}\gamma + \boldsymbol{\epsilon},
\]
but we only observe gene expression with stochasticity,
$\mathbf{m} = \mathbf{x}\alpha + \boldsymbol{\delta}$ with
$\boldsymbol{\delta}\sim \mathcal{N}\!\left(\mathbf{0}, \tau^{2}
  I\right)$.  However, we can obtain unbiased estimation of the eQTL
effect, $\boldsymbol{\mu}\equiv\mathbf{x}\alpha$, and use this to test
the mediation of this gene expression.

We set up two regression problems: (1)
$\mathbf{y} \sim \boldsymbol{\mu} \beta$ (for the mediated effect) and
(2) $\mathbf{y} \sim \mathbf{x} \theta$ (for the direct / unmediated
effect).  Straightforward algebraic derivation characterize the
distribution of estimate $\hat{\beta}$ as
\begin{eqnarray*}
  \hat{\beta} 
  &\sim& \mathcal{N}\!\left(\beta + \alpha^{-1} \gamma, \sigma^{2} n^{-1} \alpha^{-2}\right),
\end{eqnarray*}
and the estimated mediated effect size follows
\[
  \hat{\alpha}\hat{\beta}\sim\mathcal{N}\!\left(\alpha\beta+\gamma,\frac{\sigma^{2}}{n}\right).
\]

This coincides with the same distribution of direct (marginal) GWAS
statistic:
\[
  \hat{\theta} = \frac{\mathbf{x}^{\top}\mathbf{y}}{\mathbf{x}^{\top}\mathbf{x}} \sim \mathcal{N}\!\left(\alpha\beta + \gamma,  \frac{\sigma^{2}}{n}\right).
\]

\textit{Remark}: Not only we have the same mean, but also the variance
in both causal trails is symmetric. Two causal trails are
non-identifiable \cite{Gusev2016-oa,Barfield2018-hp}, unless we apply
a suitable causal inference method.

\paragraph{Asymmetric variance model}

However, if we assume stochasticity was infused before the mediation, we generate the phenotype by 
\[
  \mathbf{y} = (\mathbf{x}\alpha + \boldsymbol{\delta})\beta + \mathbf{x}\gamma + \boldsymbol{\epsilon},
\]
and it yields asymmetric variance models.  The effect size
distribution of the estimated mediated trail follows
\[
  \hat{\alpha}\hat{\beta} \sim \mathcal{N}\!\left(\alpha\beta + \gamma, \frac{\beta^{2}\tau^{2} + \sigma^{2}}{n}\right)
\]
since the gene-level association effect size follows
\begin{eqnarray*}
  \hat{\beta} 
  &\sim& \mathcal{N}\!\left(\beta + \alpha^{-1}\gamma, (\beta^{2}\tau^{2} + \sigma^{2})n^{-1}\alpha^{-2}\right).
\end{eqnarray*}

On the other hand, direct genetic association statistic takes a rather
different form of distribution:
\begin{eqnarray*}
  \hat{\theta} &=& \frac{\mathbf{x}^{\top}\mathbf{y}}{\mathbf{x}^{\top}\mathbf{x}} \sim \mathcal{N}\!\left(\alpha\beta + \gamma,  \frac{\tau^{2} + \sigma^{2}}{n}\right).
\end{eqnarray*}

{\it Remark}: These two distributions are distinguishable by the
asymmetry of variance.  Whenever there is non-zero mediation effects,
$\beta\neq{0}$, we will have larger fluctuation of the mediated
effect, but in standard error estimation we will omit
$\beta^{2}\tau^{2}/n$ and under-estimate the variance as
$\sigma^{2}/n$.  Moreover, if gene expression heritability is lower,
meaning higher $\tau^{2}$, the identification problem becomes easier.
This is somewhat paradoxical.

MR community \cite{Bowden2015-ex,Hartwig2017-qg} has adopted this type
of generative models in their simulation studies, but it appears that
we may not need strong causal assumptions to identify mediation
effects under this type of model since we can identify causal effects
by brute-force model estimation.

\subsection{Practical issues in mediation analysis}

\paragraph{Strongest correlation is not necessarily causation}

First of all, we assume true data generation scheme is much closer to
the symmetric variance model than the asymmetric variance model.
Although we have investigated the asymmetric variance assumption and
our method excels, we concluded that performance under the asymmetric
model largely depends on statistical estimation accuracy, rather than
causal inference.

\paragraph{Missing mediation problem}

All genes are heritable by definition, but only some of them are
measurable in given eQTL data. It may stem from the lack of
statistical power, or true genetic variation of expression profiles
may be conditional with respect to a certain cellular context.
Moreover, it is not difficult to imagine that a causal gene can be
included in the missing component of mediation effect.  In that case,
another non-causal gene correlated with the causal one can easily lead
to a false conclusion.


\paragraph{Pleiotropic and polygenic bias}

The missing mediation problem can be exacerbated when there is
substantial amount of unmediated genetic, thus independently
pleiotropic, effect on the phenotype within LD (the second one in
Fig.\ref{fig:causal.trails}).  It is commonly observed that this
pleiotropic effect is also highly polygenic, and creates lots of
confusions in causal mediation analysis.


\section{Causal inference on summary statistics}

\paragraph{A generative model of GWAS summary statistics}

For simplicity, letting
$S_{jj} = \hat{\sigma}_{j}^{2} + \hat{\theta}^{2}/n$, we can redefine a
model equivalent to the previous one (Eq.\ref{eq:qtl.model}) with
respect to $p$-dimensional summary statistics, the regression with
summary statistics (RSS) model \citep{Zhu2017-tl}:
\begin{equation}
  \label{eq:rss.model}
  \hat{\boldsymbol{\theta}} \sim \mathcal{N}(SRS^{-1} \boldsymbol{\theta}, SRS).
\end{equation}
Normally we have large enough sample size ($n\to\infty$), the RSS
model resorts to a fine-mapping model \citep{Hormozdiari2014-uw}.
Generative scheme of a GWAS z-score vector, with each element
$z_{j} = \hat{\theta}_{j}/\hat{\sigma}_{j}$, is described by the
reference LD matrix $R$ and true (multivariate) effect size vector
$\theta$.
\begin{equation}
  \label{eq:z.model}
  \mathbf{z} \sim \mathcal{N}(R \boldsymbol{\theta}, \sigma^{2} R).
\end{equation}

\paragraph{Reparameterized stochastic variational inference}

The key challenge in fitting the RSS model is dealing with the
covariance matrix in the likelihood. To address this challenge, we
exploit the spectral decomposition of the LD matrix
\cite{Lippert2011-lj}.  With a singular value decomposition (SVD) of
the genotype matrix, such as
\begin{equation}
  (n)^{-1/2}X = U D V^{\top},\label{eq:geno.svd}
\end{equation}
we deal with the LD matrix $R = V D^{2} V^{\top}$ and redefine a new
design matrix ${\tilde{X} \equiv V^{\top}}$ and a transformed outcome
vector
${\tilde{y}_{t} \equiv V^{\top} S^{-1}\hat{\boldsymbol{\theta}}_{t}}$
to obtain equivalent, but fully factorized, multivariate Gaussian
model:
\begin{eqnarray*}
  y_{t} \sim \mathcal{N}\!\left(D^{2}\tilde{X} S^{-1} \boldsymbol{\theta}_{t}, D^{2}\right).
\end{eqnarray*}
Further, letting
${\eta_{k} \equiv \sum_{j=1}^{p} V_{jk} S_{j}^{-1} \theta_{j}}$, we
can rewrite the transformed log-likelihood of the model for each eigen component $i$:
\begin{equation*}
\ln P(\tilde{y}_{i}|\eta_{i}) = - \frac{1}{2} \ln d_{i}^{2}
- \frac{1}{2d_{i}^{2}}(\tilde{y}_{i} - d_{i}^{2} \eta_{i})^{2} - \frac{1}{2} \ln (2\pi).
\end{equation*}
We carry out stochastic variational inference \cite{Paisley2012-fs} by
simulating stochasticity of $\eta$ by simple reparameterization
\cite{Kingma2015-mo} in the space of eigen vectors, not on the space
of SNPs in higher-dimensional space.

\paragraph{Derivation of summary-based mediation model}

We present the full description of generative model, 
\[
  \mathbf{y} \sim \mathcal{N}\!\left(\sum_{k\in[K]} X \boldsymbol{\alpha}_{k} \beta_{k} + X\boldsymbol{\gamma}, \sigma^{2}I\right),
\]
where the products of eQTL and mediation coefficients,
$\boldsymbol{\alpha}_{k}\beta_{k}$, capture mediation effects and
$\boldsymbol{\gamma}$ denotes multivariate effect sizes of the
unmediated / direct pathway.  We can reformulate an equivalent model
in terms of summary z-scores:
\begin{eqnarray*}
  \mathbf{z}^{\textsf{gwas}} &\approx& (\sigma^{2} n)^{-1/2} X^{\top} \mathbf{y} \\
  &\sim& \mathcal{N}\!\left(\frac{\sqrt{n}}{\sigma} R \sum_{k} \boldsymbol{\alpha}_{k} \beta_{k} + \frac{\sqrt{n}}{\sigma} R \boldsymbol{\gamma}, R\right).
\end{eqnarray*}
Likewise, we can characterize distribution of each gene $k$'s univariate z-score vector as:
\begin{eqnarray*}
  \mathbf{z}^{\textsf{eQTL}}_{k} &\approx& n^{-1/2} X^{\top} \mathbf{m}_{k} \\
  &\sim& \mathcal{N}\!\left(n^{-1/2} R \boldsymbol{\alpha}_{k}, \tau^{2}R\right).
\end{eqnarray*}

Assuming that we tightly controlled measurement errors in the eQTL
data, i.e., $\tau^{2} \to 0$, we can substitute the terms on the
mediation effects of the GWAS model with the z-scores of eQTL effects:
\begin{eqnarray}
  \label{eq:cammel.naive}
  \mathbf{z}^{\textsf{gwas}} &\sim& \mathcal{N}\!\left(\sum_{k\in[K]}\mathbf{z}^{\textsf{eQTL}}_{k} \beta_{k} + n^{-1/2} R \boldsymbol{\gamma}, R\right).
\end{eqnarray}

\paragraph{Identification of the unmediated ``pleiotropic'' effects}

Causality of this multivariate model can be made by statistical
inference as long as the unmediated effect $\boldsymbol{\gamma}$ is
estimable. To make it identifiable, previous methods
\cite{Barfield2018-hp,Bowden2015-ex} reduce the degree of freedom in
the $\boldsymbol{\gamma}$ parameters down to a mere intercept term.
However, our simulation suggests that sheer Bayesian inference on the
full multivariate $\boldsymbol{\gamma}$ is indeed estimable if the
GWAS and eQTL summary statistics were generated by the asymmetric
variance model.

On the other hand, in the symmetric variance model, naive inference
algorithm yields poor performance since all the genuine mediation
effects will be included in the unmediated effect.  We need to include
an additional step to construct features to characterize overall
contribution of the unmediated causal trails $X\boldsymbol{\gamma}$.

We first characterize independent components of genetic variation across 
multiple genes and diseases.  For one GWAS and $K$ eQTL z-scores, letting,
\begin{equation}
  \label{eq:z.combined}
 \tilde{Z}\equiv(\mathbf{z}^{\textsf{gwas}},
 \mathbf{z}_{1}^{\textsf{eQTL}}, \ldots,
 \mathbf{z}_{K}^{\textsf{eQTL}}),
\end{equation}
we profile overall spectrum of variation by solving the following
sparse factorization problem:
\begin{equation}
  \label{eq:factorization}
  \mathbb{E}\left[\tilde{Z}\right] = n^{-1/2} X^{\top} \left(\sum_{l\in\textrm{factors}} \mathbf{c}_{l} \boldsymbol{\omega}_{l}^{\top}\right).
\end{equation}
From this result, we obtain covariate matrix $C$, which we can
consider as projection of overall genetic variations onto reference
panel genotype space.  We use this rich vocabulary of $C$ matrix to
adjust potential unmediated effects.

However, care should be taken.  We exclude any column vector
$\mathbf{c}_{l}$ if the corresponding $\boldsymbol{\omega}_{l}$ vector
contains strong non-zero elements in both GWAS and eQTL sides.  For
instance, we call the $l$-th column is associated with gene (or trait)
$k$ if posterior inclusion probability of $\omega_{kl}$ greater than
1/2.  Our decision rule is largely compatible with the widely accepted
InSIDE (instrument strength independent of direct effect) condition
\cite{Bowden2015-ex}, but we actively search for independent
unmediated effects.  On the selected $L$ unmediated effects
$\mathbf{c}_{l}$, $l\in[L]$, we can easily construct z-scores,
$\mathbf{z}_{l}^{\textsf{unmed}} = n^{-1/2} X^{\top} \mathbf{c}_{l}$,
we then resolve mediated and unmediated effects in the following joint
model:
\begin{equation}
  \label{eq:cammel}
  \mathbf{z}^{\textsf{gwas}} \sim \mathcal{N}\!\left(\sum_{k\in[K]} \mathbf{z}_{k}^{\textsf{eQTL}} \beta_{k} +  \sum_{l\in[L]} \mathbf{z}_{l}^{\textsf{unmed}} \gamma_{l}, R\right),
\end{equation}
where both $\beta$ and $\gamma$ follow the spike-slab prior
\cite{Mitchell1988-cf}.

\paragraph{Identification of hidden non-genetic confounding effects}

The sparse factorization result (Eq.\ref{eq:factorization}) still
provides a valuable resource in checking spurious correlations
confounded by non-genetic factors (the third and fourth in
Fig\ref{fig:causal.trails}).  However, there is a risk of
over-correcting genuine genetic correlations at the same time.  We can
sidestep such a possibility by constructing a proxy data matrix
$\tilde{Z}^{(0)}$, on which we can warrant orthogonality with a
genotype matrix.  The idea is that we project our z-score matrix
$\tilde{Z}$ (Eq.\ref{eq:z.combined}) onto independent LD blocks to
adaptively construct the proxy matrix for factorization analysis.

More precisely, we define non-genetic confounding effect $\mathbf{u}$
between a gene expression $\mathbf{m}$ and phenotype vector
$\mathbf{y}$ as follows.
\begin{eqnarray*}
\mathbf{m} &=& X\boldsymbol{\alpha} + \mathbf{u} + \boldsymbol{\delta} \\
\mathbf{y} &=& X\boldsymbol{\alpha}\beta + X\boldsymbol{\gamma} + \mathbf{u} + \boldsymbol{\epsilon}.
\end{eqnarray*}
Even though we have
$\mathbb{E}\!\left[\mathbf{u}^{\top}(X\boldsymbol{\alpha})\right]=0$
by definition, gene-level correlation would have risk of including
non-causal effects:
\begin{eqnarray*}
  \mathbf{m}^{\top}\mathbf{y} = \underbrace{(\boldsymbol{\alpha}^{\top}X^{\top}X\boldsymbol{\alpha})\beta}_{\textrm{causal}} + \underbrace{\mathbf{u}^{\top}(X\boldsymbol{\gamma}) + \mathbf{u}^{\top}\mathbf{u}}_{\textrm{non-causal correlation}},
\end{eqnarray*}
where we may expect the second term to vanish with large $n$, but the
third term persists.

We propose a simple operator to make intervention only on the putative
genetic components to yield a valid proxy z-score matrix can
selectively capture non-genetic confounding effects.

As human LD patterns are close to a block-diagonal covariance matrix,
we can always find an independent LD block $X^{(0)}$ such that for all
columns $j$ of $X^{(0)}$ is orthogonal to the mediation effect, i.e.,
$\mathbb{E}\!\left[(X^{(0)}_{j})^{\top} (X\boldsymbol{\alpha})\right]
= 0$.  Before we carry out the factorization
(Eq.\ref{eq:factorization}), we project the combined z-score matrix of
$X$ onto some independent LD block $X^{(0)}$:
\begin{equation}
  \label{eq:z.proj}
  \tilde{Z}^{(0)} \gets (X^{(0)})^{\top}(X^{-\top} \tilde{Z}).
\end{equation}
As for the inverse step, we consider pseudo-inverse; by SVD
(Eq.\ref{eq:geno.svd}), $X^{-\top} = UD^{-1}V^{\top}$.  We perform
factorization on this $\tilde{Z}^{(0)}$,
\begin{equation}
  \label{eq:factorization.projected}
  \mathbb{E}\!\left[\tilde{Z}^{(0)}\right] = n^{-1/2} (X^{(0)})^{\top} C \Omega
\end{equation}
and use $n^{-1/2}X^{\top}C$ to account for non-genetic correlations.

\textit{Remark}: We can justify this can effectively eliminate genetic
effects from summary statistics: Provided that linear transformation
of multivariate Gaussian distribution yields Gaussian distribution, we
characterize the mean vector and the covariance matrix after each step
of transformation.  Without loss of generality, underlying $n$
individual-level target vector $\mathbf{y}$ has two components,
$X\boldsymbol{\alpha}$ and $\mathbf{u}$. This induces the distribution
of z-score vector:
$\mathbf{z}\sim\mathcal{N}\!\left(n^{-1/2}X^{\top}(X\boldsymbol{\alpha}
  + \mathbf{u}), n^{-1}X^{\top}X\right)$. After the first
transformation, we have
  $$X^{-\top}\mathbf{z} \sim \mathcal{N}\!\left(n^{-1/2}X\boldsymbol{\alpha} + n^{-1/2}\mathbf{u}, n^{-1} I\right).$$
  Followed by the second transformation, we have
  $$X_{0}^{\top}(X^{-\top}\mathbf{z}) \sim \mathcal{N}\!\left(n^{-1/2} X_{0}^{\top}\mathbf{u}, n^{-1}X_{0}^{\top}X_{0}\right),$$
  because $\mathbb{E}\!\left[(\mathbf{x}_{j}^{(0)})^{\top}(X\boldsymbol{\alpha})\right] = 0$ for all $j$.


\section{Experiments}

\paragraph{Simulation based on real-world genotype matrix}

To evaluate performance of our methods, we carried out extensive and
realistic sets of simulations.  Unfortunately, there is no labeled
data for causal mediation analysis; the only gold standard would be a
controlled experiment.  We might consider literature-based assessment,
but for systematic comparison, we find simulation is more adequate.

We simulate eQTL and GWAS z-scores on selected LD blocks
\cite{Berisa2016-cm} using standardized genotype matrix $X$, sampled
from the 1000 genomes reference panel
\cite{The_1000_Genomes_Project_Consortium2015-jd}, only including
individuals with European ancestry ($n$=502), and restricting on the
SNPs with minor allele frequency (MAF) $\ge$ 0.05.  This results in
the matrix $X$ ($n\times{p}$) with $n=502$ and $p$ = 5k-10k SNPs.

We have repeated our experiments using much larger cohort, such as
UK10K \citep{Huang2015-bg} samples ($n$=6,285), but results were
qualitatively identical; for brevity, we only report the results of
the 1000 genomes data.

We simulate $K=100$ gene expression vectors
$\{\mathbf{m}_{k}:k \in [K]\}$, and one phenotype vector $\mathbf{y}$.
For each simulation, we provide the following parameters:
\begin{itemize}
\item $X$: a genotype matrix (column-wise standardized).
\item $g_{g}^{2}$: proportion of gene expression variability explained
  by genetics; here, we fixed to 0.3.
\item $d$: number of causal eQTL SNPs; variability of each gene is
  determined by a linear combination of $d$ SNPs.
\item $\mathbf{u}$: in addition to genetic and unstructured noise
  components, we have unknown random effect vector.
\item $h_{m}^{2}$: proportion of phenotypic variability explained
  by genetic effects mediated through causal genes.
\item $g_{u}^{2}$: proportion of gene expression variability explained
  by the random effect $\mathbf{u}$.
\item $h_{u}^{2}$: proportion of phenotypic variability explained by
  the random effect $\mathbf{u}$.
\end{itemize}

Overall simulation steps proceed as follows.

\begin{enumerate}

\item Initially all genes are heritable.  For each gene $k\in[K]$,
  sample eQTL effect size
  $\alpha_{jk}\sim\mathcal{N}\!\left(0, g_{g}^{2}/d\right)$ for the
  causal SNP $j$ on this gene $k$, but $\alpha_{jk}=0$ for the
  others. This easily ensures
  $\mathbb{V}\!\left[X\boldsymbol{\alpha}_{k}\right] = g_{g}^{2}$.
  Genetic components of this mediator is simply
  $\mathbf{m}^{(g)}_{k} \gets X \boldsymbol{\alpha}_{k}$.

\item We follow the asymmetric variance model. Sample mediation
  effect: $\beta_{k} \sim \mathcal{N}\!\left(0, h_{m}^{2}/m\right)$
  for the causal genes, otherwise $\beta_{k}=0$; then propagate the
  mediated genetic effect to a genetic component of phenotype:
  $\mathbf{y}^{(g)} \gets \sum_{k} \mathbf{m}^{(g)}_{k} \beta_{k}$.

\item For all gene $k\in[K]$, we introduce structured random effects,
  $\mathbf{m}^{(u)}_{k} \gets \mathbf{u} \xi_{k}$ where
  $\xi_{k}\sim\mathcal{N}\!\left(0,1\right)$, and rescale this vector
  such that
  $\mathbb{V}\!\left[\mathbf{m}^{(u)}_{k}\right] = g_{u}^{2}$.

\item We do the same on the phenotype,
  $\mathbf{y}^{(u)} \gets \mathbf{u} \xi_{0}$ where
  $\xi_{0}\sim\mathcal{N}\!\left(0,1\right)$, and rescale this vector
  such that $\mathbb{V}\!\left[\mathbf{y}^{(u)}\right] = h_{u}^{2}.$

\item For non-heritable (or missing) gene $k$, we eliminate the
  genetic component, such as
  $\mathbf{m}^{(g)}_{k} \sim \mathcal{N}\!\left(\mathbf{0},
    g_{g}^{2}\right)$.  Note that this may include a causal mediation
  gene.

\item The observed expression vector on each gene $k$ is
  $\mathbf{m}_{k} \gets \mathbf{m}^{(g)}_{k} + \mathbf{m}^{(u)}_{k} +
  \boldsymbol{\delta}_{k}$ where
  $\boldsymbol{\delta}_{k} \sim \mathcal{N}\!\left(\mathbf{0},
    \tau_{0}^{2}I\right)$ with $\tau_{0}^{2} = 1 - g_{g}^{2} - g_{u}^{2}$.
\item We also observe the phenotype with the noise components:
  $\mathbf{y} \gets \mathbf{y}^{(g)} + \mathbf{y}^{(u)} +
  \boldsymbol{\epsilon}$, where
  $\boldsymbol{\epsilon}\sim\mathcal{N}\!\left(\mathbf{0},
    \sigma_{0}^{2}\right)$ with
  $\sigma_{0}^{2} = 1 - h_{m}^{2} - h_{u}^{2}$.

\end{enumerate}

\paragraph{Data} For summary statistics-based methods, we only provide
these two types of z-scores calculated from the simulated data,
$\mathbf{y}, \mathbf{m}_{k}$, \textit{not} knowing the genetic part of
data, $\mathbf{y}^{(g)}, \mathbf{m}_{k}^{(g)}$.

\paragraph{CaMMEL methods}

We term our general methodology CaMMEL (causal multivariate mediation
extended by LD) as we test multiple mediation effects simultaneously,
exploiting local LD structure.  Here, we train the CaMMEL model in
three different ways and compared them in the simulation studies:
\begin{itemize}
\item \texttt{CaMMEL-naive}: Brute-force variational Bayes inference
  of the joint model with the multivariate unmediated effect sizes
  (Eq.\ref{eq:cammel.naive}).
\item \texttt{CaMMEL-factorization}: A two-step inference algorithm where we
  first characterize the unmediated (direct) effects
  $\mathbf{z}^{\textsf{unmed}}$ by fitting the factorization model
  (Eq.\ref{eq:factorization}) and resolve the mediation effects in the
  joint modeling (Eq.\ref{eq:cammel}).
\item \texttt{CaMMEL-projection}: Another two-step inference algorithm where
  we first project the combined z-score matrix onto independent LD
  blocks (Eq.\ref{eq:z.proj}), then characterize the unmediated
  effects $\mathbf{z}^{\textsf{unmed}}$ by fitting the factorization
  model (Eq.\ref{eq:factorization.projected}) to adjust non-genetic
  / unmediated confounding effects.  Here, we adjust the GWAS z-score
  $\mathbf{z}^{\textsf{gwas}}$ by subtracting out the inferred
  $\mathbf{z}^{\textsf{unmed}}$ and resolve the mediation effects in
  the joint modeling only with the mediation terms in
  Eq.\ref{eq:cammel}.
\end{itemize}

\paragraph{Competing methods}
As for the calculation of LD-adjusted inverse-variance weighting (IVW)
and summary-based TWAS (sTWAS), we first perform SVD of the reference
genotype matrix (Eq.\ref{eq:geno.svd}), and this allows estimation of
the LD matrix by $R=VD^{2}V^{\top}$. Since we know
$\mathbf{z}\sim\mathcal{N}(R\boldsymbol{\theta},R)$. We rotate the
original distribution and define another multivariate Gaussian random
variable
$\boldsymbol{\eta} \equiv D^{-1}V\mathbf{z} \sim
\mathcal{N}(DV^{\top}\boldsymbol{\theta}, I)$ for algebraic
convenience.  Let
$\boldsymbol{\eta}^{\textsf{eQTL}}\equiv
D^{-1}V\mathbf{z}^{\textsf{eQTL}}$ and
$\boldsymbol{\eta}^{\textsf{gwas}}\equiv
D^{-1}V\mathbf{z}^{\textsf{gwas}}$.


From these two vectors, we can write sTWAS test statistics
\cite{Mancuso2017-kp}:
\[
  T^{\textsf{sTWAS}} =
  (\boldsymbol{\eta}^{\textsf{eQTL}})^{\top}\boldsymbol{\eta}^{\textsf{gwas}} / \sqrt{(\boldsymbol{\eta}^{\textsf{eQTL}})^{\top} \boldsymbol{\eta}^{\textsf{eQTL}}}.
\]
Using {\tt mr\_ivw} implemented in {\tt Mendelian Randomization} package \cite{Yavorska2017-te}, we can estimate IVW test statistics:
\[
  T^{\textsf{IVW}} = {\hat{\beta}}^{\textsf{MLE}} \max\{\hat{\sigma},1\} / \mathsf{se}(\hat{\beta}),\]
where
$\hat{\beta}^{\textsf{MLE}}, \hat{\sigma}^{2} = \arg\max
\mathcal{N}(\boldsymbol{\eta}^{\textsf{gwas}} |
\boldsymbol{\eta}^{\textsf{eQTL}} \beta, \sigma^{2}I)$ with estimated
standard error $\mathsf{se}(\beta)$. We could easily modify the IVW
method with different types of linear models, e.g., including an
intercept term in the linear model to account for directional
pleiotropy, MR-Egger regression \cite{Barfield2018-hp}.

Lastly, we compare performance with the observed TWAS (oTWAS), or
differential expression analysis, correlation between the observed
phenotype $\mathbf{y}$ and noisy observation of gene expression
$\mathbf{m}_{k}$.

\subsection{Experiments with strong polygenic bias and missing causal genes}

\begin{figure}[h]
  \centering
  \includegraphics[width=\linewidth]{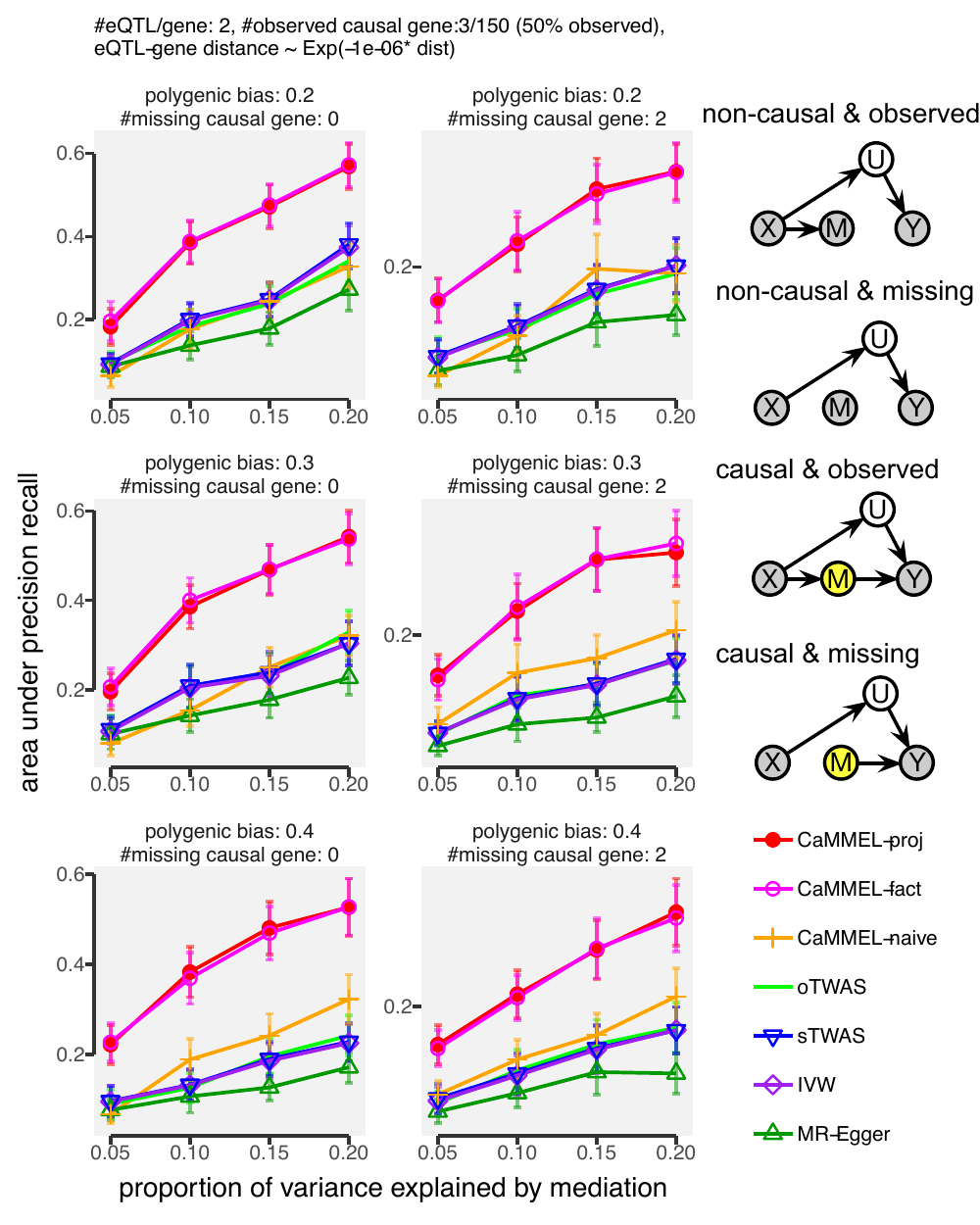}
  \caption{In the experiments under the influence of directional
    pleiotropy with 50\% of genes missing, two of the CaMMEL methods
    redeem genuine mediation trails with high accuracy.}
  \label{fig:poly.auprc}
\end{figure}

We carried out benchmark tests to evaluate robustness of causal
inference in the presence of strong polygenic bias to the phenotype
(Fig.\ref{fig:poly.auprc}).  We sampled 150 genes (using actual gene
locations in each LD block) and varied the variance of polygenic bias
($h_{u}^{2} \in \{.2, .3, .4\}$).  Of the 150 genes, we excluded 50\%
of eQTL genes in the observed statistics, which may or may not include
two causal genes.  To simulate polygenic bias, we followed a
previously suggested simulation scheme \cite{Barfield2018-hp},
$\mathbf{u} = X \boldsymbol{\gamma}$ where for each $j \in [p]$
$\gamma_{j} \sim \bar{\gamma} + \mathcal{N}(0, 10)$ with randomly
sampled direction $\bar\gamma \in \{+1, -1\}$.  We report area under
precision recall curve (AUPRC) as metric as we have much fewer causal
genes (3) compared to the non-causal ones (147). To be more exact, we
only call prediction on a gene is correct if and only if the gene is
causal and the sign of predicted effect size also matches with that of
the simulated effect.

When there is strong polygenic bias and a substantial fraction of
genes are missing, prediction accuracy (and power) of most gene
prediction methods can be severely damaged.  In human genetics data,
unmediated polygenicity is common observed across many different
traits, and it is almost impossible for us to obtain a full catalog of
eQTL genes. Interestingly, even though we did not include any
confounding effect on the mediators (genes), this type of setting is
enough to create confusion that univariate (gene-by-gene) methods to
make lots of false discoveries.  Genes are genetically dependent in LD
and become conditionally dependent given phenotype
variables. Interestingly, the MR-Egger method has been thought to
handle a directional pleiotropy (polygenic bias)
\cite{Bowden2015-ex,Barfield2018-hp}, but we could only find the worst
performance in our simulations.

On the other hand, when the existing portion of unmediated effects are
causally identified, our CaMMEL methods robustly outperform other
methods. Yet, naive inference algorithm on the CaMMEL model shows far
worse performance because the parameters on the unmediated effect
($\boldsymbol{\gamma}$) are much more adaptable to the data, and yield
far too conservative results.

\subsection{Experiments with genes and phenotypes confounded by non-genetic factors}

Next, we conducted a new type of benchmark tests where the genes and
phenotype are confounded by non-genetic effects. To set apart from the
previous experiments, we assumed genes are fully observed for
simplicity; by definition, we only considered that the non-genetic
confounders are independent of genetics.  In most mediation analysis
in genetics, we take for granted that such a confounding effect were
corrected out by pre-processing steps.  However, in practice,
especially when there were any sample overlap between eQTL and GWAS
cohorts (sharing controls), non-genetic correlations would always
exist with a high probability.  As we collect more data from biobank
(where individuals are totally shared), finding a non-genetic
confounder across multiple traits is already a crucial step in GWAS
analysis.

\begin{figure}[h]
  \centering
  \includegraphics[width=\linewidth]{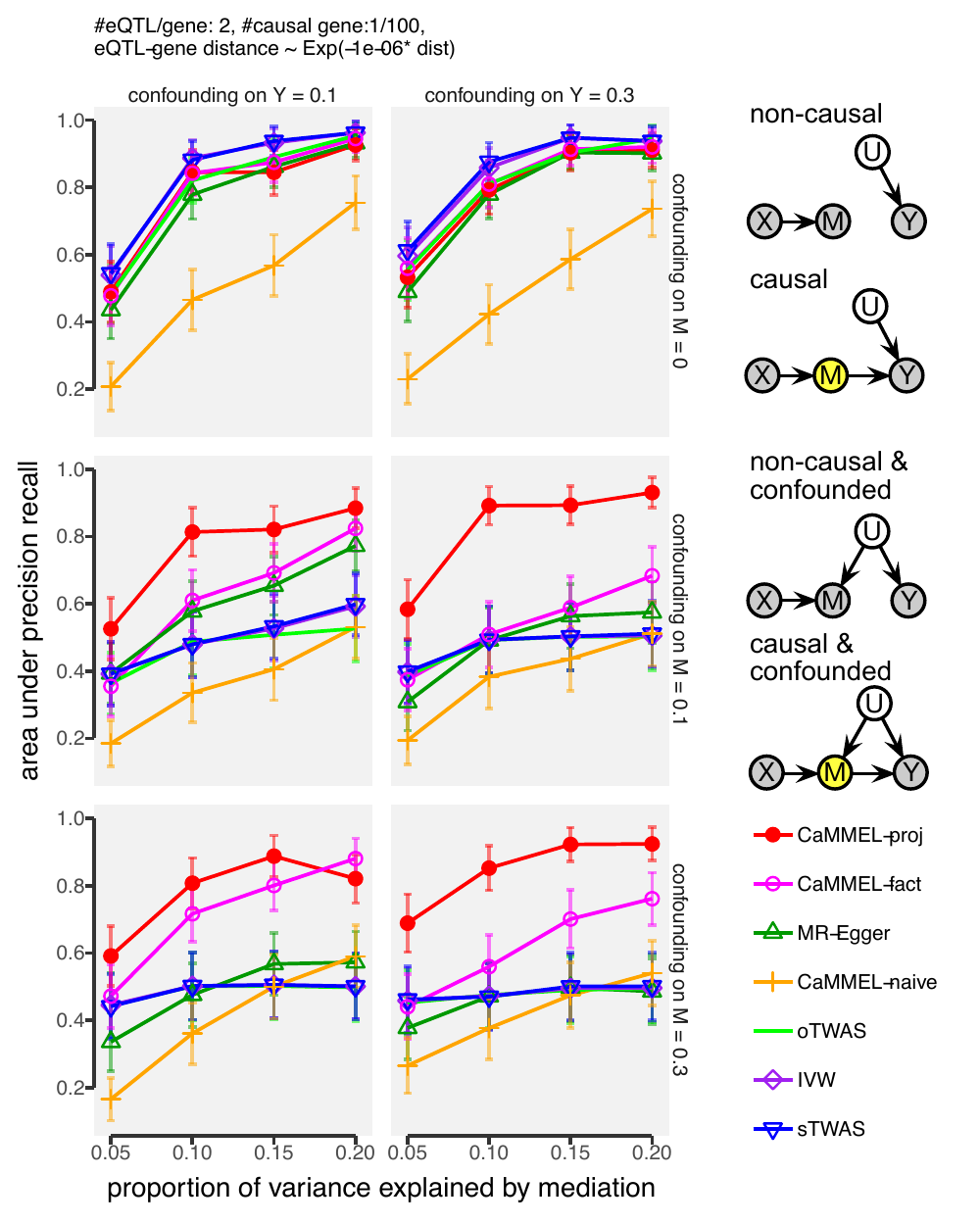}
  \caption{Our method solves a long-standing problem in GWAS:
    \texttt{CaMMEL-proj} removes non-causal confounding relationships
    between mediators and outcome variables by projecting correlation
    structure onto independent LD blocks to identify the spurious
    ones.}
  \label{fig:conf.auprc}
\end{figure}

In Fig.\ref{fig:conf.auprc}, we show results with different levels of
variability of the confounding effects on the mediator (the row
panels) and phenotype sides (the column panels). As previously, we
measure the performance in AUPRC considering that only 1 gene is
causal out of total 100 genes.  When there is no confounding (the 1st
row), all the methods, except our CaMMEL with naive inference, work
similarly, achieving nearly optimal performance.

However, we can clearly see the benefit of additional factorization
(\texttt{CaMMEL-fact}) and projection (\texttt{CaMMEL-proj}) steps in
causal inference, whenever two layers (of the mediator and outcome
variables) are confounded by unknown variables, other than
genetics. Most strikingly, our results confirm that confounding
effects become clearly separable from genetic effects in the light of
independent LD blocks (\texttt{CaMMEL-proj}); and this can be done by
a simple algebraic operation.

\section{Discussion}

The ultimate goal of mediation analysis in genetics is to impute
causality of GWAS, but previous gene-based aggregated association
methods only attempt to improve statistical power apart from causal
inference perspective.  In theory and experiments, we show that
discoveries made by a statistical method agnostic to causality can
mislead follow-up studies in practical settings.  However, our
Bayesian approach to summary-based analysis truly seeks to answer
causal questions, explicitly constructing proxy-variables to capture
the unmediated and unwanted effects.  Moreover, our framework can
robustly work against high-dimensionality and collinearity of the
parametric space, naturally induced by human genetics.

In our software (available at \url{https://ypark.github.io/zqtl}), we
not only present a specialized routine for mediation analysis, but
also provide other commonly used machine learning routines for summary
statistics analysis.  We expect much more utility in future research.

\section{Acknowledgement}
We acknowledge inspirational discussion with Liang He, Bogdan
Pasaniuc, Alkes Price, and Alexander Gusev.


\bibliography{mediation}
\bibliographystyle{icml2018}

\end{document}